\pdfoutput=1

\documentclass[11pt]{article}

\usepackage[]{acl}

\usepackage{times}
\usepackage{latexsym}
\usepackage{graphicx}
\usepackage{xurl}
\usepackage{booktabs}
\usepackage{multicol}
\usepackage{multirow}
\usepackage{listings}
\usepackage{amssymb}
\usepackage{pifont}
\newcommand{\cmark}{\ding{51}}%
\newcommand{\xmark}{\ding{55}}%
\usepackage[export]{adjustbox}
\usepackage[T1]{fontenc}

\usepackage[utf8]{inputenc}

\usepackage{microtype}

\title{CONSISTENT: Open-Ended Question Generation From News Articles}

\author{Tuhin Chakrabarty$^1$\thanks{~~Work done at The New York Times R\&D}~~~~~Justin Lewis$^{2*}$~~~~~Smaranda Muresan$^{1}$ \\
 $^1$Department of Computer Science, Columbia University\\
 $^2$The New York Times R\&D\\
 {\tt\small tuhin.chakr@cs.columbia.edu, justin@justintlewis.com, smara@cs.columbia.edu} \\
}
\begin{document}
\maketitle
\begin{abstract}
Recent work on question generation has largely focused on factoid questions 
such as \textit{who, what, where, when} about basic facts. Generating open-ended \textit{why, how, what, etc.} questions that require long-form answers have proven more difficult. To facilitate the generation of open-ended questions, we propose \textbf{CONSISTENT}, a new end-to-end system for generating open-ended questions that are answerable from and faithful to the input text. Using news articles as a trustworthy foundation for experimentation, we demonstrate our model's strength over several baselines using both automatic and human-based evaluations. We contribute an evaluation dataset of expert-generated open-ended questions.We discuss potential downstream applications for news media organizations.

\end{abstract}

\section{Introduction}

Factoid questions are relatively straightforward questions that can be answered with single words or short phrases \textit{(e.g. who, what, where, when)}. However to obtain the central idea of a long piece of text, one can ask an open-ended question \textit{(e.g. why, how, what)} \cite{cao-wang-2021-controllable,gao-etal-2022-makes}, which can essentially be viewed as an extreme summary of the text \cite{narayan-etal-2018-dont} in the form of a question. The ability to generate such questions is particularly difficult because the generated questions must be \textit{answerable} from and \textit{faithful} to the given input text (see Table \ref{table1}).

“\textit{Answer-agnostic}'' \cite{du-etal-2017-learning,subramanian-etal-2018-neural, scialom-staiano-2020-ask} or “\textit{Answer-aware}'' \cite{lewis2021paq,song-etal-2018-leveraging,zhao-etal-2018-paragraph,li-etal-2019-improving-question} question generation has 
gained focus in NLP but these approaches are usually trained by re-purposing question answering datasets that are factual in nature or trained with trivia-like factoid QA pair data sets where answers are entities or short phrases.

\begin{table}[]
\renewcommand{\arraystretch}{1.15}
\small
\begin{tabular}{|p{1.0cm}|l|}
\hline
                                                            & \begin{tabular}[c]{@{}l@{}}At the {\color{blue}current rate of COVID-19 vaccination}, \\experts say, it will take months to change the\\ virus’s trajectory. In the short term, they worry \\that the vaccine could present new risks if\\ newly immunized people start socializing \\without taking precautions. It is not yet clear if \\the vaccine protects against asymptomatic \\infection, so vaccinated people may still be able \\to spread the virus to others.\end{tabular} \\ \hline
\begin{tabular}[c]{@{}l@{}}Seq2Seq\end{tabular} & \begin{tabular}[c]{@{}l@{}}Why are people so worried about the\\ COVID-19 virus?\end{tabular}\\ \hline
\begin{tabular}[c]{@{}l@{}}Seq2Seq\\ +Control\end{tabular} & \begin{tabular}[c]{@{}l@{}}Why is the \textbf{current rate of vaccination}\\ \textbf{for COVID-19} so worrisome?\end{tabular}\\ \hline
\end{tabular}
 \caption{Example of open ended questions requiring long form answer generated by fine-tuning a Seq2Seq model BART \cite{lewis-etal-2020-bart} and by adding explicit control with salient n-grams}
 \label{table1}
\end{table}

Prior work on long-form question answering (LFQA) \cite{kwiatkowski2019natural,fan-etal-2019-eli5} focuses on generating answers to open-ended questions that require explanations. We argue that these benchmarks can also be useful for generation of diverse, human-like open-ended question requiring long form answer.

\begin{figure}[t]
    \centering
    \includegraphics[width=0.45\textwidth, frame]{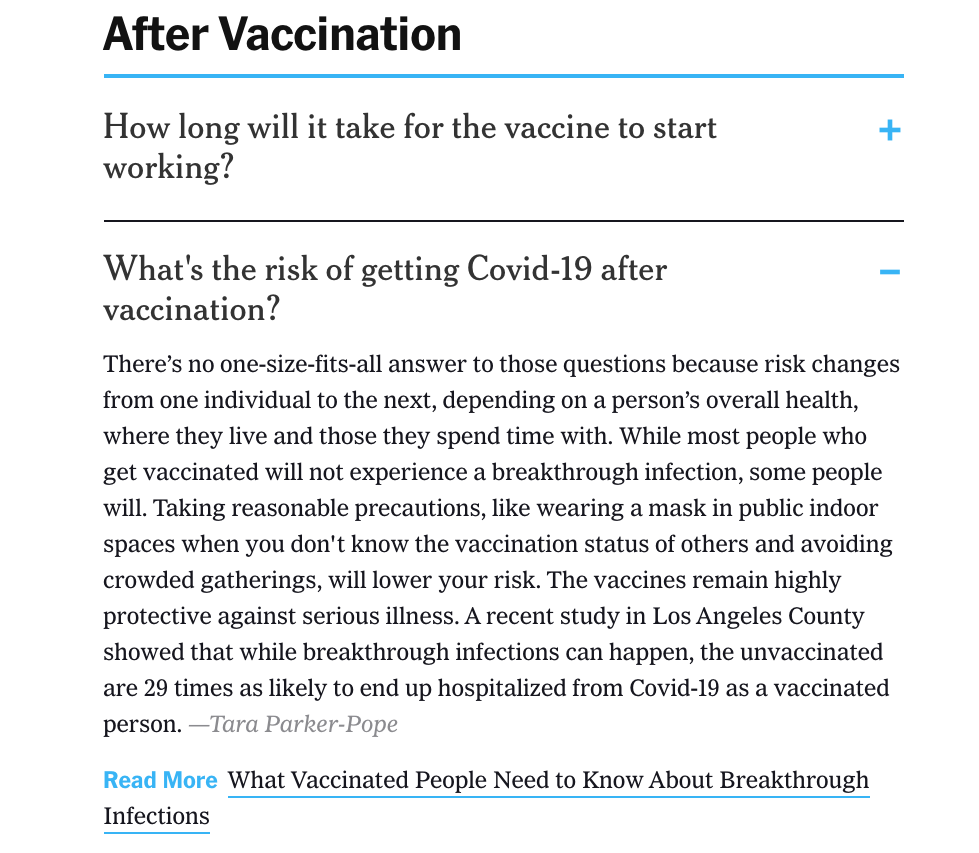}
    \caption{Human-written question-answer pairs as seen on a FAQ news tool about Covid-19 vaccination}
    \label{fig:faq}
\end{figure}

While question generation often helps in data augmentation for training models \cite{lewis2021paq,pan2020unsupervised}, 
it can also help in possible downstream consumer applications (Section \ref{downstream}). Leading news organizations often rely on human-written QA-pairs for frequently asked questions (FAQ) news tools (Figure \ref{fig:faq}) or as representative headlines for news articles used in article recommendation panels. As seen in Figure \ref{fig:faq}, a news article about \textit{the likelihood of breakthrough infections after Covid-19 vaccination} can be summarized in the form of representative question-answer pairs.

We propose a novel end-to-end system, \textbf{CONSISTENT} for generating open-ended questions that are answerable from and faithful to the input document. We fine-tune a state-of-the-art pre-trained seq2seq model \cite{lewis-etal-2020-bart} to generate open-ended questions 
conditioned on an input paragraph. We further propose methods to ensure better controllability and faithfulness for our generated questions by steering them towards salient keywords in the paragraph which act as “control codes” \cite{keskar2019ctrl}. Well-formed generated questions can still be unanswerable. Prior work on using filtering methods \cite{lewis2021paq} to ensure consistency is not possible for our task, owing to increased answer length. Thus,  
we first rely on confidence scores obtained from pre-trained question answering models to filter out simple inconsistent questions. We further evaluate answerability by designing human-readable prompts to elicit judgements for answerability from the T0pp model \cite{sanh2021multitask}, which has shown 
 good zero-shot performance on several NLP benchmarks.

We release an evaluation dataset of 529 paragraphs across diverse domains along with human written open-ended questions. Empirical evaluation using automatic metrics demonstrate that our model is better than 5 baselines. Finally, expert evaluation of the top two performing systems shows that our model is capable of generating high quality, answerable open-ended questions spanning diverse news topics (3.5 times better than a competitive baseline: a \citep[BART]{lewis-etal-2020-bart} model fine-tuned on an existing inquisitive questions-answers dataset ELI5 \citep[Explain Like I'm Five]{fan-etal-2019-eli5}. Our novel evaluation dataset, code and models is made publicly available at \footnote{\url{https://github.com/tuhinjubcse/OpenDomainQuestionGeneration}}.

\section{Related Work}
Question generation can primarily be answer-aware or answer-agnostic. Prior work on Answer-agnostic Question Generation \cite{du-etal-2017-learning,subramanian-etal-2018-neural,nakanishi-etal-2019-towards,Wang_Wei_Fan_Liu_Huang_2019,scialom-etal-2019-self} focuses on training models that can extract phrases or sentences that are question-worthy and use this information to generate better questions. \citet{scialom-staiano-2020-ask} paired questions with other sentences in the
article that do not contain the answers to generate curiosity-driven questions. However, these approaches are trained by repurposing QA datasets that are factual \cite{rajpurkar-etal-2016-squad} or conversational \cite{10.1162/tacl_a_00266,choi-etal-2018-quac}. \citet{cao-wang-2021-controllable} focus on generating open-ended questions from input consisting of multiple sentences based on a question type ontology. Most recently \citet{ko-etal-2020-inquisitive} built question generation models by fine-tuning generative language models on 19K crowd-sourced inquisitive questions from news articles. These questions are elicited from readers as they naturally read through a document sentence by sentence, are not required to be answerable from the given context or document.

Answer-Aware question generation models \cite{lewis2021paq,song-etal-2018-leveraging,zhao-etal-2018-paragraph,li-etal-2019-improving-question} typically encode a passage P and an answer A letting the decoder generate a question Q auto-regressively. These methods work well in practice and have been shown to be improve downstream QA performance. However despite their efficacy, these methods emphasize simple factoid questions whose answers are based on short and straightforward spans. Previous work on generating clarification questions \cite{rao-daume-iii-2019-answer,rao-daume-iii-2018-learning,majumder-etal-2021-ask} uses questions crawled from forums and product reviews. The answers to the questions were used in the models to improve the utility of the generated questions. 

Our work is different from prior work in that we focus on generating open-ended questions, which require long-form answers, from news articles. Unlike answer-aware question generation, where models ask a factoid question conditioned on an answer span, our task is challenging as it requires comprehension of the larger context as well as the ability to compress and represent the salient idea of the passage in the form of a question.

\section{Data} \label{data}
\paragraph{Training Data}

\begin{table}[]
\small
\renewcommand{\arraystretch}{1.25}
\centering
\begin{tabular}{|l|}
\hline
\begin{tabular}[c]{@{}l@{}}It’s springtime of the pandemic. After the trauma of the\\ last year, the quarantined are emerging into sunlight, and \\beginning to navigate travel, classrooms and restaurants.\\And they are discovering that when it comes to returning\\ to the old ways, many feel out of sorts. Do they shake \\hands? Hug? With or without a mask?\end{tabular}                                            \\ \hline
{\color{blue}How are people adapting to life after the pandemic?}                                                                                                                                                                                                                                                                                                                                                                                    \\ \hline
\end{tabular}
\caption{Examples of our evaluation data containing paragraphs from news articles with human written questions. More in Table \ref{appendixhuman} in Appendix \ref{sec:appendix}}
 \label{table2}
\end{table}
Most prior work has successfully trained models for question generation using SQUAD \cite{rajpurkar-etal-2016-squad}, TriviaQA \cite{joshi-etal-2017-triviaqa}, or NQ \cite{47761} datasets, the answers to which are typically short.

To account for the open-ended nature of our desired questions, we rely on the ELI5 \citep[Explain Like I'm Five]{fan-etal-2019-eli5} dataset. The dataset comprises 270K English-language threads in simple language from the Reddit forum of the same name\footnote{\url{https://www.reddit.com/r/explainlikeimfive/}}, i.e  easily comprehensible to someone with minimal background knowledge. 

Compared to existing datasets, ELI5 comprises diverse questions requiring long-form answers. It contains a significant number of open-ended \textit{how}/\textit{why} questions. Interestingly, even \textit{what} questions tend to require paragraph-length explanations (\textit{What is the difference...}). As seen in Table \ref{tableeli} in Appendix \ref{sec:appendix}, each question is open-ended, inquisitive and requires an answer that is descriptive in nature. Finally, one of the advantages of the ELI5 dataset is that it covers diverse domains such as \textit{science, health, and politics}. This quality makes ELI5 an ideal candidate to transfer to the news domain, which similarly covers a diverse range of topics.

\paragraph{Evaluation Data}
Since our goal is to generate open-ended questions from news articles, we specifically design our evaluation data to reflect the same. To achieve this goal we obtain English-language articles from \textit{The New York Times} website from January 2020 to June 2020. We obtained written consent to use this content for research purposes by the copyright holder. One of the additional advantages of crawling data from the \textit{The New York Times} website is that we can divide news articles by domain, as each news article appears in a specific section of the website. From the given URL\footnote{ \url{https://www.nytimes.com/2021/12/10/science/astronaut-wings-faa-bezos-musk.html}}, we can tell that the article belongs to the \textit{Science} domain. Additionally, as most pre-trained language models were trained prior to the Covid-19 pandemic, we also test how well they generalize to COVID-19 related news topics.

Each news article from a particular domain is segmented into several paragraphs. We randomly sample \textbf{529} paragraphs spanning six domains. This includes 55 paragraphs from Science, 66 from Climate, 98 from Technology, 110 from Health, 100 from NYRegion, and 100 from Business. While we understand that selecting standalone paragraphs might sometimes ignore the greater context, or suffer from co-reference issues, we carefully replace any such paragraphs from our bigger pool.

As we do not have gold questions associated with each paragraph, we crowd-source human-written questions for each paragraph on Amazon Mechanical Turk. Each paragraph is shown to a distinct crowdworker who is then instructed to read the paragraph carefully and write an open-ended question that is answered by the entire passage. We recruit 96 distinct crowd workers for this task. After the questions are collected from first round of crowd-sourcing, two expert news media employees approve or reject them based on quality. The paragraphs with rejected questions are put up again and through this iterative process and careful quality control we obtain one high quality open-ended question associated with each paragraph. Table \ref{table2} and \ref{appendixhuman} shows selected paragraphs from our evaluation set and the associated human-generated open-ended question.

\section{CONSISTENT Model}  \label{consistent}

\begin{figure}[t]
    \centering
    \includegraphics[width=0.5\textwidth]{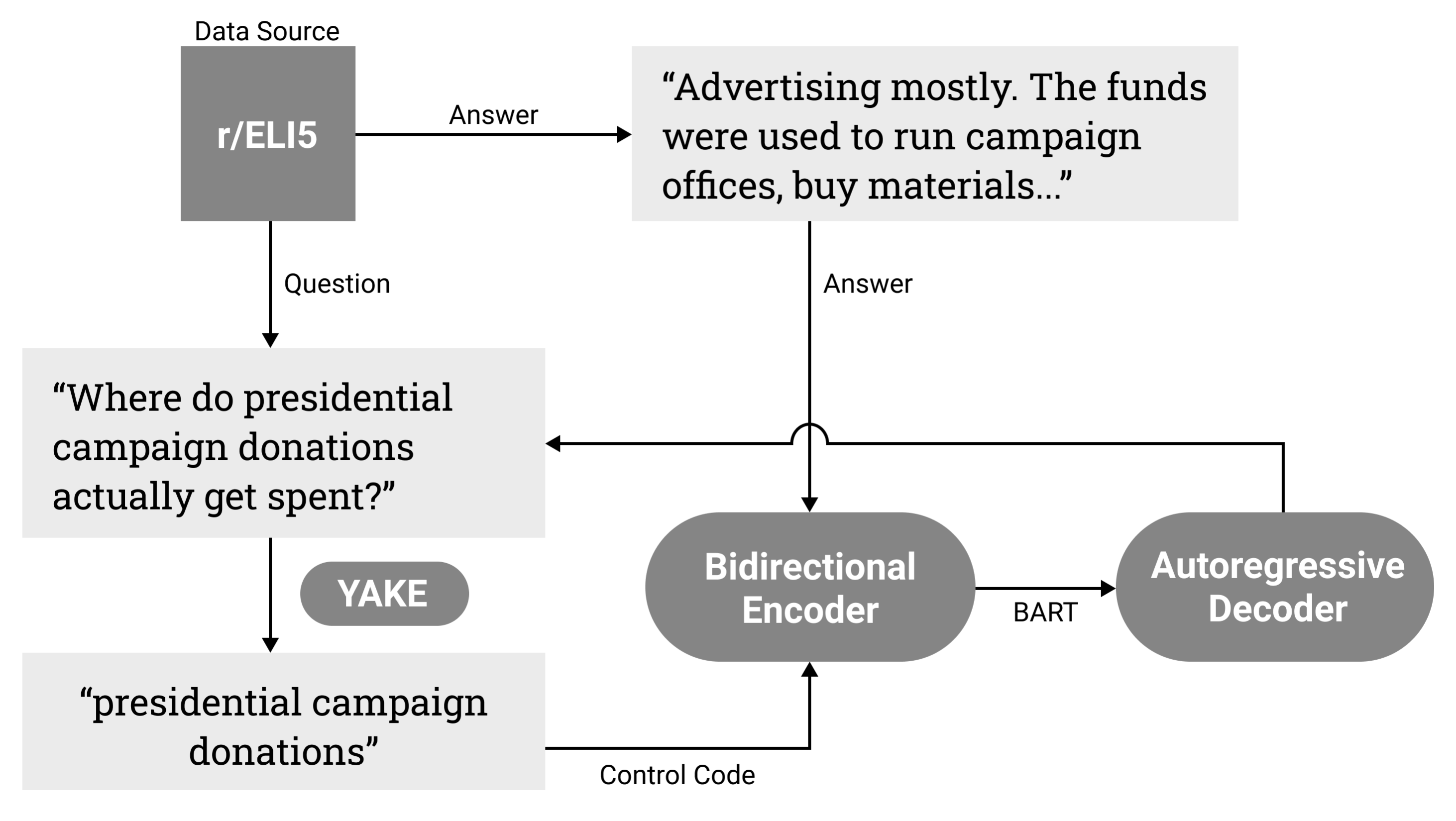}
    \caption{Architecture to train the CONSISTENT model}
    \label{fig:training1}
\end{figure}

\begin{figure*}[t]
    \centering
    \includegraphics[width=0.9\textwidth,height=0.5\textwidth]{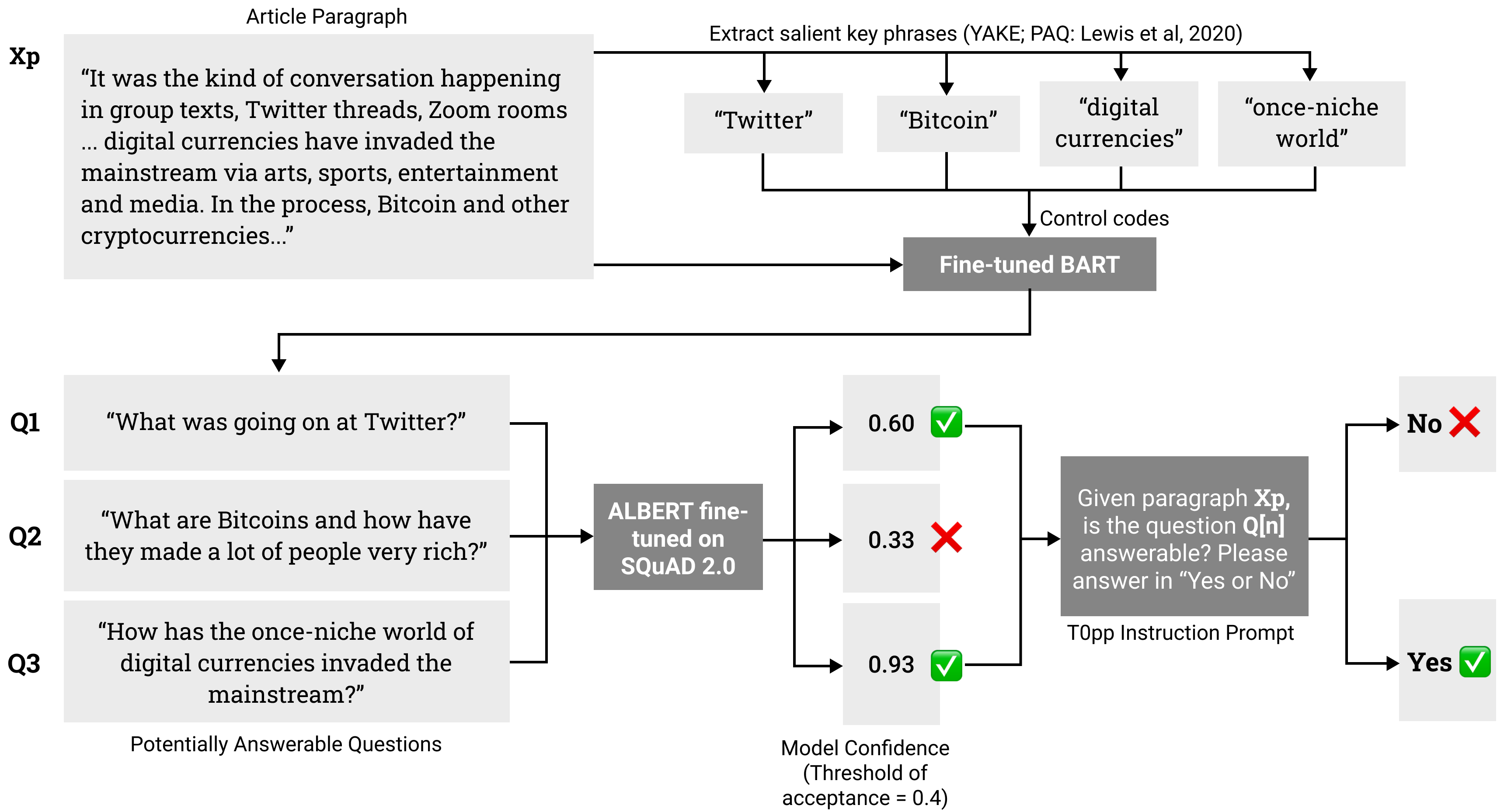}
    \caption{Inference pipeline for CONSISTENT model ensuring Factual Consitency (Control Codes) and Answerability (Model Confidence and Instruction prompting)}
    \label{fig:inference1}
\end{figure*}

The backbone of our approach is a fine-tuned BART-large \cite{lewis-etal-2020-bart} model on the ELI5 dataset of question-answer pairs. However, there are two major factors to consider in our end-to-end question generation pipeline. The generated questions i) must be \textit{relevant and factually consistent} to the input paragraph, and ii) must have the \textit{answer self-contained} in the input paragraph. Our CONSISTENT model (Figures  \ref{fig:training1} and \ref{fig:inference1}) addresses these issues as described below.

\paragraph{Factual Consistency}
To ensure faithfulness to the input paragraph, we need to design our model in such a way that the generated question is about a topic or concept mentioned in the paragraph. In traditional fine-tuning of a seq2seq model where \textit{x} denotes input paragraphs in the training set and \textit{y} denotes the corresponding question our goal is to learn $p_{\theta}(y|x)$ where
\begin{equation}
        p_{\theta}(y|x) = \prod_{i=1}^{n} p_{\theta}(y_{i}|y_{i-1},y_{i-2}.....y_{1},x)
\end{equation}

Recently \citet{keskar2019ctrl} proposed CTRL, a conditional language model that is conditioned on a control code \textit{c} and learns the
distribution $p_{\theta}(y|x,c)$ to provide explicit control over text generation. The distribution can still be decomposed using the chain rule of probability and trained with a loss that takes the control code into account.
\begin{equation}
        p_{\theta}(y|x,c) = \prod_{i=1}^{n} p_{\theta}(y_{i}|y_{i-1},y_{i-2}.....y_{1},x,c)
\end{equation}

Owing to this modification, language models can generate text conditioned on control codes that specify domain, style, topics, dates, entities, relationships between entities, plot points, and task-related behavior. We rely on the same underlying principle for training question generation models.

During training, we extract keywords from questions and feed the input paragraph along with the extracted keyword to the encoder of BART. The extracted keyword here acts as the control code. Since we do not have any supervision for these keywords we use YAKE \cite{Campos2020YAKEKE}, an unsupervised keyword extraction tool. For example, as shown in Figure \ref{fig:training1}, given the question: \textit{Where do presidential campaign donations actually get spent?}, we extract the top-most salient trigram \textit{“presidential campaign donations”} using YAKE. We then feed the  (control code, answer) to the encoder, the original question to the decoder, and fine-tune the model as shown in Figure \ref{fig:training1}.

\citet{lewis2021paq} propose a BERT \cite{devlin2018bert} based answer extraction model on Natural Questions (NQ) by predicting $p(a | c) = p([a_{start}, a_{end}] | c)$ where “a” is an answer and “c” is a passage containing “a”. This model first feeds the passage “c” through BERT, before concatenating the start and end token representations of all possible spans of up to length 30, and then feeds them into an MLP to compute $p(a|c)$. At generation time, the answer extraction component extracts a constant number of spans from each passage, ranked by their extraction probabilities. These extracted spans, while originally designed for a different purpose, can act here as control codes for our question generation model. To encourage the question to refer to a concept mentioned in the passage, we extract salient key phrases as control codes from the input paragraph using a combination of YAKE and the answer extraction model $p(a|c)$ (Figure \ref{fig:inference1}).

It should be noted that during training the keywords are taken from the question, while during inference the keywords are produced from the article. This is because at training time we want to minimize the generation loss with respect to the training question so encouraging the model to obey the training keyword is beneficial. At inference time we do not have access to any question so using keywords from article is the only option.

\paragraph{Answerability}
Prior work \cite{lewis2021paq,fang2020accelerating,alberti-etal-2019-synthetic} has relied on filtering methods to ensure answerability of generated questions. A filtering QA model $p_f(a|q,C)$ generates an answer for a given question. If an answer generated by $p_f$ does not match the answer a question was generated from, the question is discarded. Such filtering methods are not applicable for our task because i) our question generation model treats the entire input paragraph as an answer instead of short answer spans typically common in ODQA tasks, and ii) the length of the answers are typically long-ranging across several sentences which is beyond the capabilities of most generative models \cite{krishna2021aurko} and additionally would be hard for string matching purposes. We propose two filtering methods to ensure answerability: model confidence and instruction prompting.

{\bf Primary Filtering: Model Confidence}
QA models trained on SQUAD 2.0 \cite{rajpurkar-etal-2018-know} are capable of asserting when a question is unanswerable by signaling lower confidence scores. Taking advantage of this fact we first rely on an ALBERT-based QA model finetuned on SQUAD 2.0 \footnote{\url{https://huggingface.co/mfeb/albert-xxlarge-v2-squad2}}. The intuition behind this is such a model would typically have lower confidence scores for most poorly formed / unanswerable questions and can be used as a primary filtering step.

While it may appear that a model trained on SQuAD to determine the answerability of questions conflicts with open-ended nature of questions requiring long-form answers, it is often not the case. As we have seen in the case of Natural Questions \cite{kwiatkowski2019natural} at least 35\% question requiring a long form answer often has a short answer associated with it. This means the specific span that is returned by the SQUAD based QA model when prompted with the generated question and the input paragraph can often be an approximate  short answer. For instance for the generated question Q3 in Figure \ref{fig:inference1}, the SQUAD model gives an answer \textit{via art, sports, entertainment and media.} which isn't  inaccurate but requires further elucidation. This motivates us to use a model trained on SQUAD v2.0 as our initial primary step.
However different questions can have different model confidence. To decide on an appropriate threshold for model confidence we observe the distribution of confidence scores. We observe a median model confidence of $0.42$. We then experiment with 3 different thresholds $\kappa  \epsilon  \{0.35,0.4,0.45\}$ for selecting generated questions. The quality on a held out set of 50 generated questions by is evaluated by human judges and finally decide on a model confidence threshold $\kappa=0.4$ such that any generated question having a confidence score below $\kappa$ is discarded. It should also be noted that we tried higher values of $\kappa$ between (0.6,0.9) but having such a strict high confidence score sometimes leaves us with no generated question for an input paragraph. As can be seen in Figure \ref{fig:inference1}, a generated question “\textit{What are Bitcoins and how have the made a lot of people very rich?}” while being open-ended, grammatically correct, and relevant to the input does not meet the answerability threshold and hence is discarded.

{\bf Secondary Filtering: Instruction Prompting}
While the above filtering step acts as excellent proxy for unanswerable questions, the original model is still trained for short answer spans. To ensure our filtering method is devoid of such biases, we use a secondary filtering approach. Recently \citet{sanh2021multitask} show how large language models exhibit zero-shot generalization to unseen tasks when presented with natural language prompts. As we do not have annotated data for answerability judgements for open-ended questions with longer answer spans, we rely on zero-shot prompt-based instructions for further filtering. We prompt the best-performing model from \citet{sanh2021multitask} \textit{T0pp}  with the following instruction:

\begin{lstlisting}[backgroundcolor=\color{gray!20!white},showstringspaces=true,breaklines=true,basicstyle=\ttfamily,xleftmargin=\parindent]
Given paragraph <@\textcolor{blue}{\{\{paragraph\}\}}@>,
is the question <@\textcolor{blue}{\{\{question\}\}}@>
answerable? Please answer 
in Yes or No
\end{lstlisting}

\begin{table}[]
\vspace{-.9ex}
\small
\renewcommand{\arraystretch}{1.25}
\begin{tabular}{|p{1.55cm}|p{5.35cm}|}
\hline
Input  & \begin{tabular}[c]{@{}l@{}}The variant from South Africa, known as\\ B.1.351, could make things even worse for\\ the vaccine push. Given the speed at \\ which the variant swept through that\\ country, it is conceivable that by April it\\ could make up a large fraction of infections\\ in the United States.\end{tabular}                                    \\ \hline\hline

BART   & What's going on with the Ebola virus?                                                                                             \\ \hline
Lead   & What is the name of the variant from South Africa?                                                                                             \\ \hline
SQUAD   & What is B.1.351?                                                                                             \\ \hline
RandomOut   & Why is the domestic product of the flu so bad right now?                                                                                             \\ \hline
RandomIn & \begin{tabular}[c]{@{}l@{}}What is going on in the US right now after a\\ B.1 variant swept through the country?\end{tabular}\\ \hline
CONSISTENT & \begin{tabular}[c]{@{}l@{}}What does the variant from South Africa\\ mean for the vaccine push?\end{tabular}                                                                                                    \\ \hline
\end{tabular}
\caption{\label{table:examples}Generated Questions from Baseline Models and CONSISTENT.
}
\end{table}

We feed the questions that pass the acceptability test based on our model confidence threshold as natural language instructions to the model as shown in Figure \ref{fig:inference1}. Only questions which receive an answer of “Yes” are considered in our final set. This process makes our filtering approach robust, owing to the fact that only questions which pass both filtering tests are considered as \textit{consistent}. 

It can be argued that T0pp is a stochastic system that was not trained for un-answerability detection. To justify our use of T0pp we conducted a experiment where we sample a subset of 200 questions (100 answerable and 100 unanswerable). These questions are manually selected by humans from our pool of all possible generated questions. We then feed T0pp with the same prompt above containing the respective questions and their associated paragraphs. On a binary task of un-answerability prediction we get an accuracy of 84\%.

As our pipeline can generate multiple questions for each input due to different control codes, we further need to rank the generated questions. Towards this task, we rank all our generated questions for a given input paragraph that are consistent based on model confidence scores. 

\section{Evaluation Setup} \label{evaluation-section}

\subsection{Baselines}
We compare our CONSISTENT model against several baseline approaches. 

\paragraph{Lead Sentence to Question (Lead):} In order to ensure that our data is free from any potential artifacts we take the lead sentence of every passage and convert it to a question. In particular, we prompt the T0pp \cite{sanh2021multitask} model which acts as a statement-to-question converter transforming the first sentence of every paragraph to a question.

\paragraph{QG based on fine-tuned BART (BART):} Our initial backbone model of  fine-tuned BART-large on answer-question pairs from the ELI5 dataset.

\paragraph{QG based on random keyword inside Paragraph (RandomIn):} We use the same fine-tuned BART-large model from Section \ref{consistent} with <keyphrase, paragraph> as input to the encoder and the question as the output from the decoder. During inference we feed a random keyphrase from the input paragraph  to generate the question. It should be noted that this approach does not undergo any of the filtering mechanism used in CONSISTENT.

\paragraph{QG based on random keyword outside Paragraph (RandomOut):} The training method is similar to that of RandomIn except that during inference we feed a random keyphrase outside of the input paragraph to generate the question. It again does not undergo any of the filtering mechanism used in CONSISTENT.

\paragraph{QG based on SQUAD data (SQUAD):} We fine-tuned a BART-large model on SQUAD 2.0 but conditioning on the keyphrases in the prompt during inference. In particular, we use the same keyphrase used in the prompt for the highest scoring question from our CONSISTENT model. For instance, we prompt the model fine-tuned on SQUAD with the keyphrase \textit{once-niche world} and the input paragraph as shown in Figure \ref{fig:inference1}.

\subsection{Evaluation Metrics}

The space of possible correct outputs is too large in our case to rely on n-gram based metrics like BLEU or ROUGE. For this reason, we chose the two best available automatic evaluation metrics based on contextual representations. We report BERTScore \cite{Zhang-etal:2020:bertscore} to measure the similarity between a generated question and its gold-reference human written question.\footnote{We used BERTScore based on \textit{deberta-mnli} that is shown to have high correlation with human judgements.}. We also report BLEURT \cite{sellam-etal-2020-bleurt} scores, which combine expressivity and robustness by pre-training a fully learned metric on large amounts of synthetic data, before fine-tuning it on human ratings.
\begin{table}[]
\renewcommand{\arraystretch}{1.15}
\small
\centering
\begin{tabular}{|l|c|c|}
\hline
           & BLEURT & BERTScore   \\ \hline
BART       & 44.0 & 64.5 \\ \hline
Lead       & 43.0 & 64.4 \\ \hline
SQUAD      & 39.0 & 62.1 \\ \hline
RandomInside     & 40.0 & 62.1 \\ \hline
RandomOutside     & 34.2 & 58.1 \\ \hline\hline
CONSISTENT & \bf{47.0*} & \bf{66.4*} \\ \hline
\end{tabular}
\caption{\label{autoeval}Evaluation based on automatic metrics. *Results are significant ($p<0.005$) via t-test.}
\end{table}

However, automatic metrics are not enough. To evaluate the controllability and answerability of the generated open-ended questions we chose outputs from 2 best performing systems based on the automatic evaluation. We further propose a new metric \textit{well-formedness} and a human-based evaluation. A \textit{well-formed} question is grammatically correct, faithful to the provided paragraph, and whose answer is detailed, long-form spanning through the entire paragraph. A well-formed question only mentions people, places, things, or ideas that are also in the original text.

Regarding human judges, \citet{karpinska-etal-2021-perils} discuss how even with strict qualification filters, AMT workers are not suitable for evaluating open-domain NLG outputs. To avoid such issues we recruit multiple employees of a news media organization with experience in building products and tools for news room to evaluate the output of our baseline and CONSISTENT model for each input paragraph. We believe these evaluators can ground their judgments in the real-world utility of the generated questions for our target use case. Each input was evaluated by three people. Annotator guidelines are in Appendix \ref{sec:appendix}. We use the Amazon SageMaker Ground Truth\footnote{\url{https://aws.amazon.com/sagemaker/data-labeling}} platform where we upload our input paragraphs along with the generated questions from two systems (randomly shuffled) as shown in Figure \ref{fig:evaluation}. The news article headline is given for additional context to the human evaluator. The evaluators are provided with the above definition of what constitutes a \textit{well-formed} question. The evaluators are then asked to determine which questions are well-formed between four possible options Question1, Question2, Both Questions, and Neither Question. 

\begin{table}[]
\small
\renewcommand{\arraystretch}{1.15}
\begin{tabular}{|p{1.3cm}|p{1.8cm}|p{1.15cm}|p{0.6cm}|p{0.65cm}|}
\hline
           & \textbf{CONSISTENT} & \textbf{BART} & \textbf{Both} & \textbf{None} \\ \hline
Health     & \textbf{56.3}       & 13.6     & 17.2  & 12.7    \\ \hline
Technology & \textbf{37.7}        & 22.4    & 22.4  & 17.3     \\ \hline
\begin{tabular}[c]{@{}l@{}}Science \&\\ Climate\end{tabular} & \textbf{42.1}      & 18.1     & 23.1 & 16.5   \\ \hline
New York &  \textbf{51.0}       &  9.0   & 33.0  &  7.0   \\ \hline
Business &   \textbf{55.0}      &  11.0   & 30.0 &   4.0  \\ \hline\hline
\textbf{Overall} &   \textbf{48.4}      &  14.8   & 25.1  &  11.5    \\ \hline
\end{tabular}
\caption{Human-based evaluation results (percentage win).
$p<.0001$ via t-test}
\label{tab:results}
\end{table}

\section{Results and Analysis}
Table \ref{autoeval} shows that our CONSISTENT model is better than all the existing baselines. Table \ref{tab:results} shows that our experts agree on the quality of the generated questions spanning different domains. To get a single verdict on the correct label for each input, we consider majority voting for each question. Agreement rates were measured using Krippendorff’s $\alpha$ and a moderate agreement of 0.62 was achieved. As observed, our CONSISTENT model outperforms Baseline BART overall by a margin of 33.6 points. Table \ref{table:examples} shows examples of generations by the five models on a given paragraph. In an effort to better understand why or how the CONSISTENT model is better than the baselines, we carefully analyze outputs from all systems.

Without an explicit supervision on what to ask, BART often asks generic questions or deviates from the input source and hallucinates content as can be seen in Table \ref{table:examples} and Table \ref{hallucination}. The LEAD model works decently well when the central idea of the paragraph is expressed in the first sentence, however without broader context it often suffers from generating factoid or uninteresting questions. The SQUAD model is the second-worst performing model as expected due to mismatch in training and evaluation domain. Due to lack of answerability filters, the RandomIn model even though generating a question based on a random keyword from article is often found to be unanswerable as demonstrated by the automatic evaluation scores in Table \ref{autoeval}. The worst performance of RandomOut model bolsters our claim that using keywords from the article in the prompt helps the model achieve faithfulness and doing otherwise might hurt performance.

To test the effect of the primary filtering we choose the candidate question with the highest confidence given by the QA model. To test the effect of secondary filtering we choose the candidate question with the highest confidence of generating yes. The obtained BERTScore for Primary, Secondary and CONSISTENT(both) are 64.0, 64.5 and 66.4 showing that the best filtering mechanism is the combined one.

\section{Downstream Applications} \label{downstream}

\begin{figure}[t]
    \centering
    \includegraphics[width=0.48\textwidth,frame]{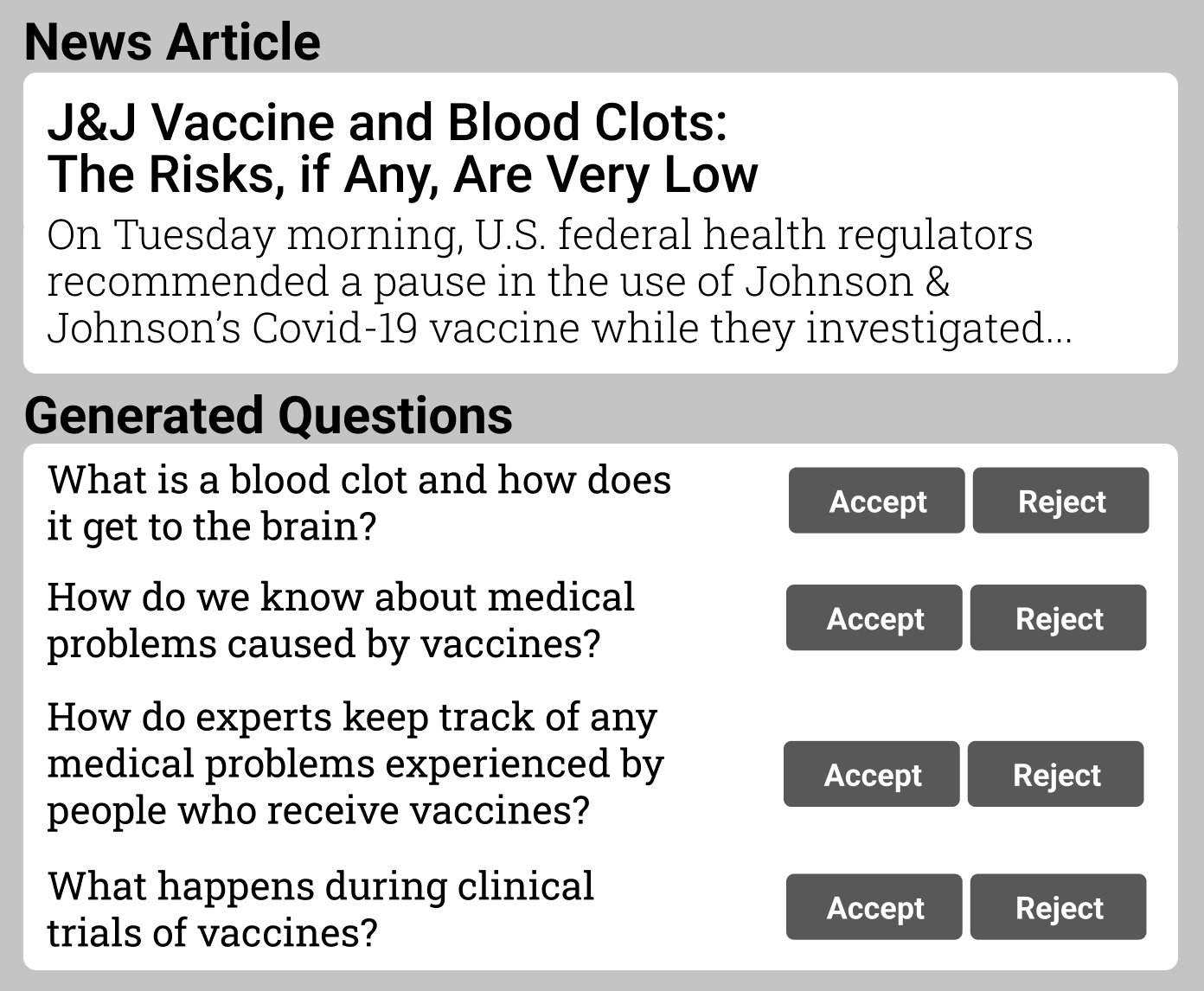}
    \caption{A prototype admin tool for humans to approve, reject, or edit questions generated by CONSISTENT for individual news articles.}
    \label{fig:humanadmin}
\end{figure}

We believe open-ended question generation might enhance the news experience through new Q\&A tools, enhanced search, improved recommendations, and more. Media organizations have used FAQ pages to help readers better understand complex news topics, from Covid-19 vaccines\footnote{\url{https://www.nytimes.com/interactive/2021/well/covid-vaccine-questions.html}} to personal finance\footnote{\url{https://www.washingtonpost.com/business/2021/12/07/faq-new-debt-collection-rules/}}. The ability to automatically generate open-ended question about a given topic could make it easier for news organizations to launch an FAQ page for a new topic. We envision an admin tool (Figure \ref{fig:humanadmin}) that presents the users with a list of generated questions and allows them to approve/reject, edit, and publish the results quickly. This human-in-the-loop approach is essential for maintaining reader trust when the generated questions may be presented directly to readers \cite{laban-etal-2022-quiz}.

\begin{figure}[t]
    \centering
    \includegraphics[width=0.48\textwidth,frame]{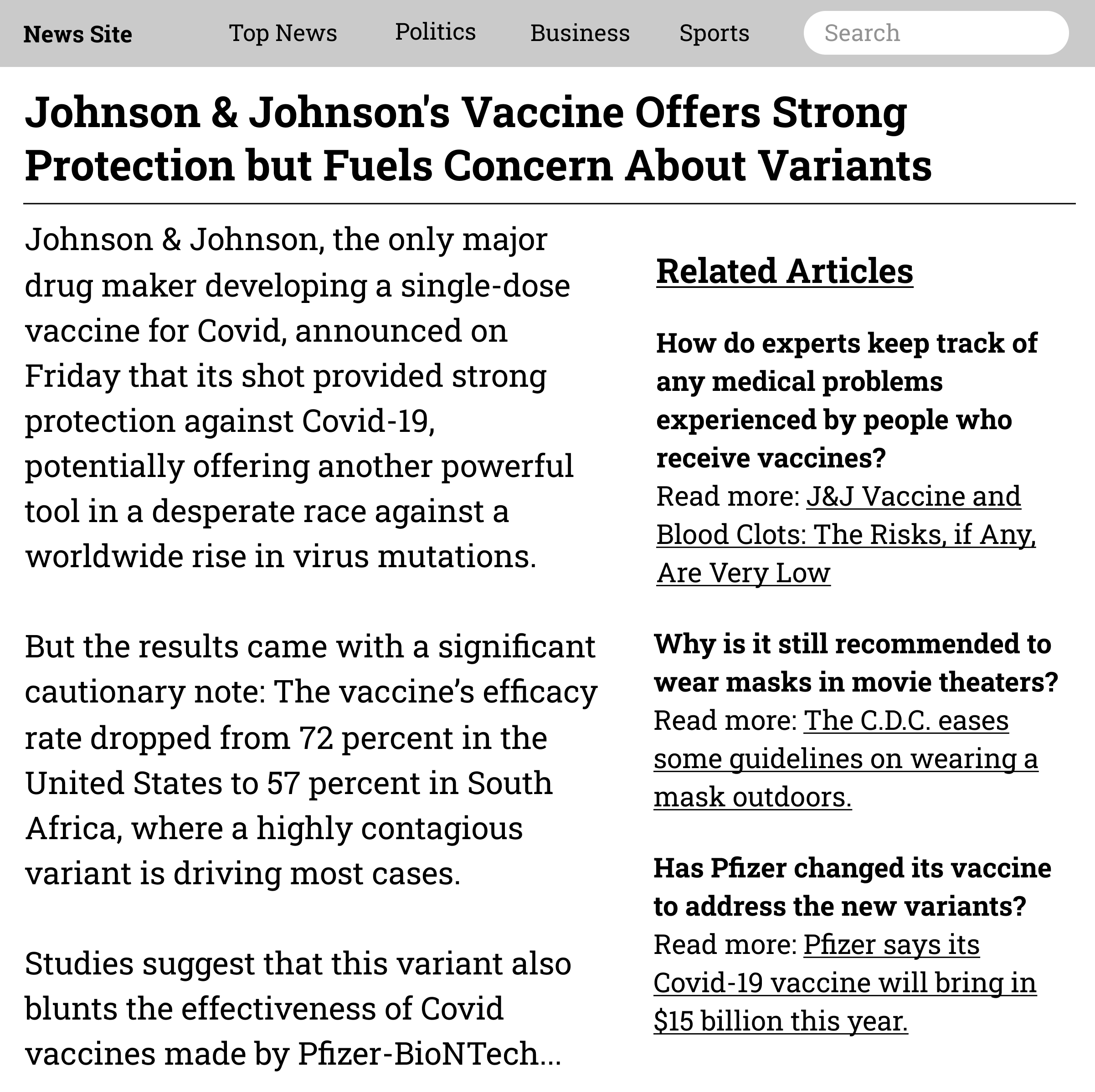}
    \caption{A prototype news article where human-approved questions generated by CONSISTENT are displayed as related articles.}
    \label{fig:related}
\end{figure}
Another potential application of this human-in-the-loop version of the system is improved news article recommendations (Figure \ref{fig:related}). While reading a news article, a recommended article interface may be presented showing a series of questions related to the topics in the article. This level of functionality could in some ways anticipate questions a reader may ask and point them in the direction of other news articles that may provide them with answers.

For other use cases that might allow automatically generated open-ended question to be used more broadly in production systems, we envision a human-validated database of such question-answer pairs where the reliability of the results could be controlled. This can help improve the search experience on news media websites. For instance, if a user were to search \textit{What are some of the issues with NFTs?}, the search experience could fuzzily match questions generated by CONSISTED to prioritize the article containing the answer relevant to the user's query which can help them discover things pertaining to what they are curious about. Additionally, question clustering algorithms could be deployed to better match searched questions with the generated questions. Finally automatic question generation can also used to improve interactive news podcast \citet{laban2022newspod}.

\section{Conclusion}
We propose CONSISTENT, an end-to-end system for generating open-ended questions  requiring long form answers, 
which accounts for factual consistency and answerability. Using news articles as a trustworthy foundation for experimentation, we demonstrate CONSISTENT’s strength over a competitive baseline model as evaluated both using automatic metrics and human evaluation. We also contribute an evaluation set of input paragraphs and human-generated open-ended questions. Through potential downstream applications of CONSISTENT, we demonstrate how they can enhance the experience of news media websites.

\paragraph{Ethical Considerations}

As noted in Section \ref{data}, we use a corpus of news articles from \textit{The New York Times} as the foundational set of documents used for question generation. We have used this data with the approval and consent of the copyright holders for research purposes. We intentionally decided to use news articles as a trustworthy foundation for question generation. Further, we selected \textit{The New York Times} as they have published\footnote{\url{https://www.nytco.com/company/standards-ethics/}} a clear set of ethics and standards to guide the creation of their journalism.Our models were trained on four A100 GPUs for 10 hours. Parameter size 400m.

As with other text generative models, our model can suffer from hallucinations \cite{hallucination}, biases \cite{sheng-etal-2019-woman, sheng2021societal} from the Reddit ELI5 dataset and text found on the internet more broadly, and concerns about potential misuse. Much of the paper goes into detail about the great lengths we have gone to in order to reduce hallucinations and exert greater control over the final outputs in order to counter these risks (see Section \ref{consistent}). We use control codes selected from the original news article in an attempt to better control the generated question. We filter for answerability to further ensure that generated questions are faithful to the original text. While considerable work has been done to reduce the impact of these issues, any language generation system will be imperfect.

Our human evaluators were selected due to their familiarity with standards of journalism. Each evaluator was a paid, full-time employee of a news media organization.

To encourage critical thinking about the risks of deployment in a production environment, we included Section \ref{downstream} to discuss possible downstream applications. We detailed our perspective on when a human-in-the-loop would be essential to an ethical use of this system. 

We hope that our work in this paper can further the important work of safe and trustworthy language generation.

\section{Limitations}

We note that our training dataset is automatically collected from the r/ELI5 subreddit and as such we don't account for any sensitive text. We focus on open ended question generation from news articles where our inputs are paragraph level and longer than sentence level inputs in factoid QG. However our model is not capable of handling longer sequences like an entire news article or opinion piece. We believe models like LongT5 \cite{guo-etal-2022-longt5} might be useful for such inputs however we leave this for future task.

Even though we control for hallucination by incorporating control codes from input text, it does not ensure 100\% hallucination free output. In regards to answerability judgements our methods are useful and bridge the gap in distinguishing unanswerable questions however it in itself is a difficult task and our approaches based on SQUAD V2.0 and T0pp can still make errors.This means our models are still capable of generating unacceptable questions and should be deployed based on due deliberation.

Finally temporal misalignment is an issue and owing to the fact that our training data is from a few years back it sometimes fails on newly coined scientific terms or expressions related to COVID-19 pandemic. Continually fine-tuning our models on newer data with experience replay can mitigate these issues. We leave this for future work.

\section*{Acknowledgements}
We would like to thank the anonymous reviewers for their helpful comments. Tuhin was funded by NYC Media Lab x New York Times R\&D Researcher Fellowship. The authors also want to thank the members of NYTimes R\&D team and NYC Media Lab teammates Robert Clauser, Erica Matsumoto and Matt Macvey for their support. 

\bibliography{anthology,custom}
\bibliographystyle{acl_natbib}

\appendix
\section{Appendices}
\label{sec:appendix}
\paragraph{Annotator Guidelines}

As our well-formedness metric constitutes multiple dimensions it is important for us to have clear annotation guidelines. Towards this we specifically instruct workers on what should be looked into

\begin{itemize}
    \item Question needs to be grammatical
    \item Question should not refer to concepts or entities that is not referenced in the original paragraph. For instance, a question about an \textit{Ebola vaccine} when the original text is about the \textit{COVID-19 vaccine} is NOT considered well-formed. Also a question that references \textit{Vice President Biden} when the text is about \textit{President Biden} would not be considered well-formed.  
    \item Question needs to be faithful and relevant to the input paragraph and on same topic
    \item The question should encapsulate or summarize the key idea of the entire passage and should not be simply factoid (i.e something that can be answered using a few words) 
\end{itemize}

\begin{table}[h]
\renewcommand{\arraystretch}{1.4}
\small
\begin{tabular}{|l|l|}
\hline
 & \begin{tabular}[c]{@{}l@{}}The variant from South Africa, known as B.1.351, \\ could make things even worse for the vaccine push.\\ Given the speed at which the variant swept through\\ that country, it is conceivable that by April it could\\ make up a large fraction of infections in the United\\ States.\end{tabular} \\ \hline
 {\color{red}\xmark}     & What's going on with the Ebola virus?                                                                                                                                                                                                                                                                                         \\ \hline
Type  & Hallucination                                                                                                                                                                                                                                                                                                                 \\ \hline
 {\color{red}\xmark}     & What is the name of the variant from South Africa?                                                                                                                                                                                                                                                                            \\ \hline
Type  & Simply factoid                                                                                                                                                                                                                                                                                                                \\ \hline
 {\color{red}\xmark}     & What is B.1.351?                                                                                                                                                                                                                                                                                                              \\ \hline
Type  & Simply factoid                                                                                                                                                                                                                                                                                                                \\ \hline
{\color{green}\cmark}      & \begin{tabular}[c]{@{}l@{}}What does the variant from South Africa\\ mean for the vaccine push?\end{tabular}                                                                                                                                                                                                                  \\ \hline
Type  & Well-formed                                                                                                                                                                                                                                                                                                                   \\ \hline
\end{tabular}
\caption{\label{annotation}Generated examples used to instruct human workers as to what makes a question well-formed and why are some questions not well-formed}
\end{table}

They were also provided with examples of generated questions along with reasoning so as to why a given question is not well-formed.

\paragraph{Hyperparameters}
We fine-tune a BART Large model on ELI5 for both baseline and CONSISTENT for 10 epochs with batch size 64 and learning rate $5e-6$ and save the best checkpoint based on validation loss. To generate questions we use top-k sampling \cite{fan2018hierarchical} with $k=5$ and a temperature of 0.8 coupled with $no\_repeat\_ngram\_size=2$

\begin{table}[]
\small
\renewcommand{\arraystretch}{1.15}
\begin{tabular}{|l|l|}
\hline
         Input & \begin{tabular}[c]{@{}l@{}}At the two New Horizons homes in\\ Gainesville, the medical director, Dr. Swati\\ Gaur has held six staff town halls, in person \\or online, including one at 2 a.m. for the \\night shift, and offered rewards like free \\meals. About half of the workers have been \\vaccinated, Dr. Gaur said.\end{tabular} \\ \hline
BART & \begin{tabular}[c]{@{}l@{}}{\color{red}How do nursing homes deal with the}\\ {\color{red}influx of new residents?}\end{tabular}                                                                                            \\ \hline
Consistent & \begin{tabular}[c]{@{}l@{}}What has the medical director of New\\ Horizons done to ensure that workers \\ are vaccinated?\end{tabular}                                                                     \\ \hline
\end{tabular}
\caption{\label{hallucination}Hallucination by the Baseline BART model}
\end{table}

\begin{table*}[!h]
\renewcommand{\arraystretch}{1.15}
\small
\centering
\begin{tabular}{|l|l|}
\hline
Question                                                                                        & Answer                                                                                                                                                                                                                                                                                                                                                                                                                                                                                                                                  \\ \hline
\begin{tabular}[c]{@{}l@{}}Why are my muscles sore \\ after jumping in cold water?\end{tabular} & \begin{tabular}[c]{@{}l@{}}From what I understand, our bodies defenses against hypothermia is to shiver.\\ This involves involuntary muscle contractions to generate heat. These muscles \\ contractions still can cause muscle soreness just like working out.\end{tabular}                                                                                                                                                                                                                                                  \\
 \hline
\begin{tabular}[c]{@{}l@{}}How come bluetooth is \\  so much slower than Wi-Fi?\end{tabular}              & \begin{tabular}[c]{@{}l@{}}Bluetooth is designed to be short-range very low-power for small portable \\ equipment. Part of the power-savings of Bluetooth come from diminished bandwidth \\ (just as much as the weaker signal). One could speed up Bluetooth to Wi-Fi speeds, \\ but then it would defeat the purpose of BT's major design feature. If you're looking\\for something that works like plunging a cable between devices but has Wi-Fi\\speeds, you might like wireless USB.\end{tabular}                                                               

\\ \hline
\end{tabular}
 \caption{Examples from the r/ELI5 Subreddit of open ended question requiring long form answer paired with human-written answers}
 \label{tableeli}
\end{table*}

\begin{figure*}[t]
    \centering
    \includegraphics[width=0.9\textwidth,height=0.5\textwidth, frame]{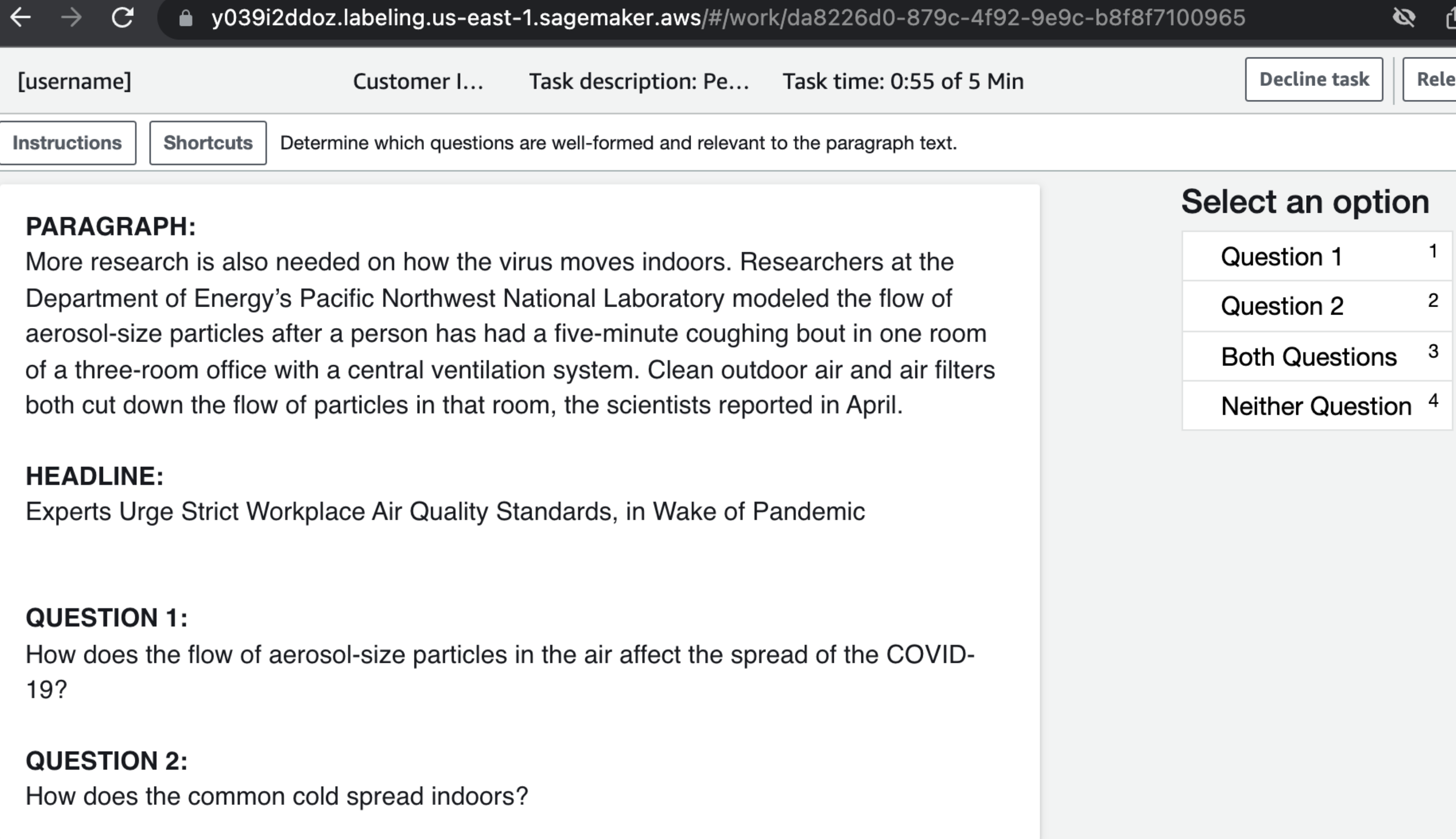}
    \caption{Screenshot of the evaluation tool where employees of a news media organization are asked to select the best option given two generated questions, an input paragraph, and headline from a news article}
    \label{fig:evaluation}
\end{figure*}

\begin{table}[]
\small
\renewcommand{\arraystretch}{1.25}
\centering
\begin{tabular}{|l|}
\hline
\begin{tabular}[c]{@{}l@{}}More companies are also using augmented reality to \\help people with online shopping, Ms. Ask said. Jins\\ Eyewear, which sells prescription glasses, lets you take a \\photo of your face to virtually try on glasses before \\deciding whether to buy them. Snap, the parent company\\ for Snapchat, has teamed up with luxury brands like \\Gucci and Dior to offer virtual try-ons.\end{tabular} \\ \\ \hline
{\color{blue}How are companies using AR for online shopping?}\\ \hline
\begin{tabular}[c]{@{}l@{}}\\For instance, a number of U.S. colleges and universities, \\including the University of Arizona and the University \\of North Carolina at Charlotte, have used wastewater \\surveillance of dorms to find asymptomatic, infected \\students who had otherwise evaded detection. In the \\Netherlands, health officials have used wastewater data\\ to determine where to send their mobile testing buses,\\ Dr. Medema said.\end{tabular} \\ \hline
{\color{blue}How has wastewater data been used to detect symptoms?}\\ \hline
\begin{tabular}[c]{@{}l@{}} \\No matter what their goals are — moving a stock, overturning\\ a presidential election, getting the graphics on a Sonic the\\ Hedgehog movie changed — these internet-based insurgencies\\ tend to follow a similar pattern. One day, a group decides\\ to take action against a system it feels is immoral or corrupt.\\ Members identify structural weak points (a vulnerable\\ political party, a risk-averse studio head, an overexposed\\ short position) and figure out creative ways to exploit\\ them, using social media for leverage and visibility. With\\ enough highly motivated people pushing in the same \\direction they eventually prevail, or get enough attention that \\it feels like they did.\end{tabular} \\ \hline
{\color{blue}How do internet-based insurgencies gain traction?}\\ \hline

\begin{tabular}[c]{@{}l@{}} \\A growing body of research shows that FEMA often helps \\white disaster victims more than people of color, even when \\the amount of damage is comparable. The problem seems to \\stem from complex systemic factors, like the difficulty of \\navigating the federal bureaucracy and a real estate market that \\often places higher values on properties in communities\\ with white residents.\end{tabular} \\\\ \hline
{\color{blue}Why does FEMA serve more white victims?}\\ \hline

\begin{tabular}[c]{@{}l@{}} \\The demands come as the safety of firefighters has become\\ an urgent concern amid worsening effects of climate change,\\ which bring rising temperatures that prime the nation for \\increasingly devastating fires. In October,2 dozen firefighters \\in California where a record 4.2 million acres burned across the \\state last year — filed suit against 3M, Chemours, E.I. du Pont \\de Nemours and other manufacturers, claiming that the \\companies for decades knowingly made and sold firefighting\\ equipment loaded with toxic chemicals without warning of \\the chemicals’ risks.
.\end{tabular} \\ \hline
{\color{blue}Why are firefighters suing companies in California}\\ \hline

\end{tabular}
\caption{\label{appendixhuman}Human written questions from our crowdsourced evaluation set}
\end{table}

\end{document}